\documentclass[lettersize,journal]{IEEEtran}
\usepackage{amsmath,amsfonts, amssymb}
\usepackage{algorithmic}
\usepackage{algorithm}
\usepackage{array}
\usepackage{textcomp}
\usepackage{stfloats}
\usepackage{url}
\usepackage{verbatim}
\usepackage{graphicx}
\usepackage{cite}
\usepackage{hyperref}
\usepackage{subcaption}
\usepackage{xcolor}
\usepackage{balance}
\usepackage{booktabs}
\begin{document}


\title{Is That Rain? Understanding Effects on Visual Odometry Performance for Autonomous UAVs and Efficient DNN-based Rain Classification at the Edge}

\author{Andrea Albanese, Yanran Wang, Davide Brunelli, David Boyle
\thanks{Corresponding author: Andrea Albanese, email: andrea.albanese@unitn.it}
\thanks{Andrea Albanese (andrea.albanese@unitn.it) and Davide Brunelli (davide.brunelli@unint.it) are with the Department of Industrial Engineering at the University of Trento, Italy.}
\thanks{Yanran Wang (yanran.wang20@imperial.ac.uk) and David Boyle (david.boyle@imperial.ac.uk) are with the Systems and Algorithms Lab at Imperial College London, United Kingdom.} \thanks{The source code is available at the following link: \url{https://github.com/albaHC/RainSeverityClassification}}
}

\markboth{Journal of \LaTeX\ Class Files,~Vol.~14, No.~8, August~2021}%
{Shell \MakeLowercase{\textit{et al.}}: A Sample Article Using IEEEtran.cls for IEEE Journals}


\maketitle

\begin{abstract}

The development of safe and reliable autonomous unmanned aerial vehicles relies on the ability of the system to recognise and adapt to changes in the local environment based on sensor inputs. State-of-the-art local tracking and trajectory planning are typically performed using camera sensor input to the flight control algorithm, but the extent to which environmental disturbances like rain affect the performance of these systems is largely unknown. In this paper, we first describe the development of an open dataset comprising $\sim$335k images to examine these effects for seven different classes of precipitation conditions and show that a worst-case average tracking error of 1.5 m is possible for a state-of-the-art visual odometry system (VINS-Fusion). We then use the dataset to train a set of deep neural network models suited to mobile and constrained deployment scenarios to determine the extent to which it may be possible to efficiently and accurately classify these `rainy' conditions. The most lightweight of these models (MobileNetV3 small) can achieve an accuracy of 90\% with a memory footprint of just 1.28 MB and a frame rate of 93 FPS, which is suitable for deployment in resource-constrained and latency-sensitive systems. We demonstrate a classification latency in the order of milliseconds using typical flight computer hardware. Accordingly, such a model can feed into the disturbance estimation component of an autonomous flight controller. In addition, data from unmanned aerial vehicles with the ability to accurately determine environmental conditions in real time may contribute to developing more granular timely localised weather forecasting.

\end{abstract}


\section{Introduction}

Autonomous Unmanned Aerial Vehicles (UAVs) are set to become central to a variety of industrial applications ranging from first response to infrastructure communications to deliveries, among myriad other UAV-based Internet of Things (IoT) services~\cite{7572034}. In each case, UAVs will require the ability to safely navigate under variable weather conditions. Although much recent research attention has been paid to autonomously navigating complex environments characterised by the presence of obstacles and dynamic disturbances from airflow~\cite{Jung18,Tijtgat17}, little effort has been afforded to understanding the effects of other dynamic environmental factors, particularly rain, as also highlighted in recent articles~\cite{Halford24,Luo21}.
Visual odometry (VO) and visual inertial odometry (VIO) leveraging depth cameras are one of the most promising methods to achieve autonomous navigation~\cite{8600375, 8967651}, however, little attention has been paid to understanding the effects of rainy conditions in terms of tracking errors that might be expected when the camera lens is exposed to various rainy conditions. 


Rainfall can be expected to disrupt the visual scene by altering image contrast, introducing blurring effects, and potentially obscuring visual landmarks or obstacles due to water droplets on the camera lens. These factors can reasonably be expected to lead to significant degradation in VO performance, resulting in inaccurate position and motion estimation for the UAV. In worst-case scenarios, inaccurate navigation due to rain could lead to mission failure, collisions, and other safety hazards. Thus, developing methods to identify and estimate the severity of rain conditions impacting VO accuracy is critical to ensure safe and reliable UAV navigation in variable weather.

The most relevant contributions in the literature concerning this or similar problems have emerged from the autonomous driving point of view~\cite{9817108, ZHANG2023146, tan2023localization}. In these cases, the images used in the analysis and estimation of the environmental conditions are taken from inside the cockpit; therefore, the water droplets are on the windshield and not directly on the camera lens. On the other hand, such contributions cannot be directly applied to developing reliable autonomous UAV applications because, in such scenarios, the UAVs navigating in rainy conditions may often have the camera lens directly exposed to the environment. 
While some examples of datasets collected with exposed camera lenses exist, e.g.,~\cite{Burnett23} and \cite{Wenzel20}, these are also in the context of autonomous driving where the instances of `rain' or `rainy' conditions are limited, and the lack of systematic collection and labeling granularity (i.e., within class) renders them insufficient for our purposes. As a result, we are motivated to study and analyse the effects of rain on a VO system suitable for autonomous UAVs with direct lens exposure. Given the absence of a suitable relevant dataset, we designed a set of laboratory experiments to simulate a flight at a low altitude (i.e., a scene with objects) under a variety of rainy conditions to collect and label with a view to identifying and classifying precipitation conditions in real time.

Moreover, many authors have successfully used deep learning (DL) or deep neural networks (DNN) to predict and estimate rain severity in vehicles~\cite{wang2021deep, kondapalli2021real}. This serves as inspiration to leverage such algorithms to estimate rain conditions in order to improve our system's performance and reliability. However, autonomous vehicles can have relatively large computational resources, while small drones (i.e., typically carrying a payload up to 2 kg) have necessarily limited computational resources considering size and payload capacity. Thus, when designing a DL-based system, we must remain aware of the available onboard resources.

This paper presents a first step towards the development of lightweight models that can determine precipitation conditions in real time, and which are suitable for use in future autonomous flight controller designs. We begin by determining the extent to which various intensities of rain can introduce tracking errors to VO-based navigation systems. We do this leveraging a large dataset that we have collected under controlled laboratory conditions. A low and fixed altitude flight scenario is developed, where a depth camera, processing unit and mechanical spraying apparatus are used to simulate various rain conditions. A dataset comprising the images taken for various rain intensities and orientations was curated and used as a basis to develop a DNN-based system to classify and estimate the severity of the rain in each case. 


DNN training has been performed with a view to ensuring a low-complexity algorithm that can be deployed on edge (e.g., small drones) or IoT-type devices~\cite{9778241}. By accurately estimating rain severity, the system can provide a basis for the development and implementation of appropriate counteractions (i.e., control strategies) to maintain reliable navigation performance during UAV operations under dynamic rainy conditions. 

We summarise the main contributions of this paper as:
\begin{itemize}

    \item We have collected and made openly available a new dataset\footnote{The dataset ($\sim$12 GB compressed) is available at:~\url{https://ieee-dataport.org/documents/adverse-rainy-conditions-autonomous-uavs}} comprising approximately 335k real images equally distributed among 7 classes that represent different levels of rain intensity, spanning clear to slanting heavy rain. Unlike other similar datasets comprising images taken from the cockpit, the camera lens is directly exposed to the water droplets in our case. 
    
    \item We provide the characterization of a VO system under different rain intensities in order to demonstrate the varying consequences to tracking accuracy, obtaining an average error in path estimation ranging from 0.07 m to 2.5 m. This permits quantifying the average error and recovery time, and may serve as a basis for designing suitable strategies to safely navigate in dynamic rainy weather conditions.

    \item We present the training, testing, and comparison of three state-of-the-art DNNs to evaluate the feasibility of the approach and analyze the performance and resource requirements. 
    Our results demonstrate that various off-the-shelf DNN-architectures developed for mobile or constrained processors offer excellent low-latency classification accuracy under almost all conditions. The top-performing model is MobileNetV3 Small achieves 90\% accuracy with a frame rate of 93 FPS and a memory footprint of just 1.28 MB.
    
\end{itemize}

The paper is organized as follows: Section~\ref{SOTA} analyses the related work concerning UAV navigation in adverse conditions. Section~\ref{expsetup} presents the experimental setup used to conduct the experiments, and Section~\ref{VOresults} summarizes the related results. Section~\ref{DNN} presents the DNN used and the training setup, while Section~\ref{DNNtest} presents the different DNN test results. Finally, Section~\ref{conclusion} concludes the paper with a view to future work.

\section{Related Works}
\label{SOTA}

UAV deployment is increasing rapidly thanks to commercial devices that are easily accessible to professionals, researchers and amateur enthusiasts alike. Despite recreational use, UAVs are excellent tools to support a variety of industrial application scenarios. In future, autonomous UAVs are likely to be able to navigate in a coordinated `swarm' where each UAVs is a node or agent of an IoT system~\cite{9321460}. In this setting, they can collaborate to exchange data, obtain a more accurate data collection, and efficiently complete critical mission tasks~\cite{9779853, 9504602}. They offer advantages in extreme environments where human intervention may be hazardous. Moreover, they avoid the need for a specifically trained and certified pilot to control them on an individual basis, consequently, increasing their reliability and opening their usage to many applications~\cite{albanese2022low, albanese2022design, santoro2023plug}.
A major constraint, however, continues to be the requirement to operate safely in highly dynamic outdoor environments mostly affected by weather. Most contemporary UAV systems cannot fly in all weather conditions, limiting their usage for time-limited missions (e.g., search and rescue), and under the control of human pilots. 

The authors in~\cite{gao2021weather} have studied UAV  ``flyability", which is ``the proportion of time drones can fly safely". On average, a common drone has a `flyability' lower than 5.7 h/day (or 2.0 h/day considering only daylight hours). However, this estimate does not consider all weather conditions, especially extreme ones, such as slanting heavy rain or high-speed wind. This analysis suggests increasing the drone's weather resistance to improve its flyability. For instance, a weather-resistant drone may increase its flyability to 20.4 h/day (or 12.3 h/day considering only daylight hours). This research confirms the fundamental role of weather in drone navigation. However, it does not take into account the drone's autonomous navigation, thus the perturbation of the sensing and navigation systems involved in this technology. 

Researchers are studying autonomous navigation systems for UAVs with deep-reinforcement learning to increase their reliability as components of Internet of Things systems~\cite{10129124, 8993742}. However, there is a knowledge gap and missing contributions that demonstrate the effects of variable weather conditions on these autonomous UAV systems. Many researchers have begun to study these effects from an autonomous driving point of view. Accordingly, the sensing systems and data involved are tailored for autonomous vehicles, and are thus not directly applicable to small autonomous UAVs. For instance, datasets of images available to the research community in the context of autonomous vehicles are taken from within the vehicle cockpit and looking out through the windshield. This is a setting that may not be similar for UAVs~\cite{RobotCarDatasetIJRR, RCDRTKArXiv}, where camera lenses are often directly exposed to the environment. Nonetheless, these works show the effect of adverse weather conditions in autonomous vehicles, suggesting clearly that similar conditions will be present and affect autonomous UAVs. We therefore expect that this can be studied and addressed by adopting similar methodologies~\cite{fursa2021worsening, hnewa2020object, appiah2024object}.

Adverse weather conditions comprise perturbed situations that are caused by, e.g., wind, rain, snow, fog, and flares. For example, the authors in~\cite{wang2023trustworthy, wang2022kinojgm, wangprobabilistic} have studied autonomous UAV navigation under perturbations caused by external airflow or wind. In particular, they have proposed a solution based on reinforcement learning (RL) to tackle unknown external disturbances whilst guaranteeing exponential convergence for any feasible reference trajectories~\cite{wang2023quadue}. Such works represent valuable initial research in this field as they open the usage of autonomous UAVs under external wind, but leave open opportunities to examine and solve for the effects of additional environmental disturbances.

Considering the variety of under-explored adverse weather conditions, we focus on rain given that it is one of the most common variable conditions, and can be expected to be responsible for significant perturbation to VO systems and consequently damaging to the accuracy and safety of local trajectory planning. Accordingly, we focus on the camera system responsible for the sensor data provided as input to the VO algorithm. Raindrops can bring different perturbations to the sensing and perception system, such as refraction or reflection of light rays. This results in pixel value fluctuation, thus causing the inaccurate processing of the raw images. For instance, DL-based algorithms (e.g., object detection and image classification) are heavily affected by rain, making them unreliable in such conditions. Thus, it is important to be aware of the navigation condition to optimize the involved image processing and computer vision algorithms~\cite{brophy2023review}.
Of course, the performance achieved with visual odometry methods can be enhanced via sensor fusion, which may include wireless-inertial methods to cope with extreme weather conditions, low light, or camera saturation. For example,~\cite{Zhang21} demonstrates how two coupled modules with RF angle-of-arrival (AoA) positioning can estimate UAV attitudes, fused with odometry measurements, to optimize vehicle poses.

Furthermore, researchers have proposed de-raining methods as image post-processing techniques to mitigate the rain effects. However, such techniques are likely to be minimally effective in our case given performance limitations and their relatively large computational complexity~\cite{zhang2023data, yan2023image, zhang2022gtav}. Meanwhile, other contributions focus on the implementation of specific algorithms that can outperform standard ones in adverse conditions, although these solutions may not be translatable as they are tailored to deal with specific applications, and thus their utility in other scenarios is questionable~\cite{appiah2024object, huang2022sfa}.

In general, the impact and severity of variable weather, particularly rain, is understudied. As a consequence, we believe that this research may be important in the development and implementation of fully autonomous UAVs that can safely navigate outdoors in all conditions.

\section{Experimental Setup} \label{expsetup}
Our initial objective is to determine the extent to which various rain conditions affect the performance of a state-of-the-art VO system used for autonomous local trajectory tracking and generation. We make the assumption that depth perception (or other) cameras mounted on UAVs are likely to have the camera lens directly exposed to the environment. This follows the majority of related literature on autonomous UAV systems that leverage VO and VIO for trajectory tracking and generation. As such, in addition to specifying the sensor and computer architecture (Sec.~\ref{hw}), some mechanical design to ensure ingress protection against moisture is a prerequisite (Sec.~\ref{IP}).

\subsection{Hardware Specification}\label{hw}
The key sensor and computer hardware underpinning the VO system comprise a processing unit, i.e., an Intel NUC 11~\footnote{Specifically, we use Intel NUC 11 Pro Kit NUC11TNKi7 as the computation unit.}, and a depth camera, i.e., an Intel Real Sense D435i. These are typical components in use among researchers developing autonomous UAV systems leveraging visual odomoetry~\cite{qin2017vins,wang2022kinojgm}.  

\subsection{Mechanical Design for Moisture Protection}\label{IP}
\begin{figure}[t!]
\centering
\subfloat{\includegraphics[width=40mm]{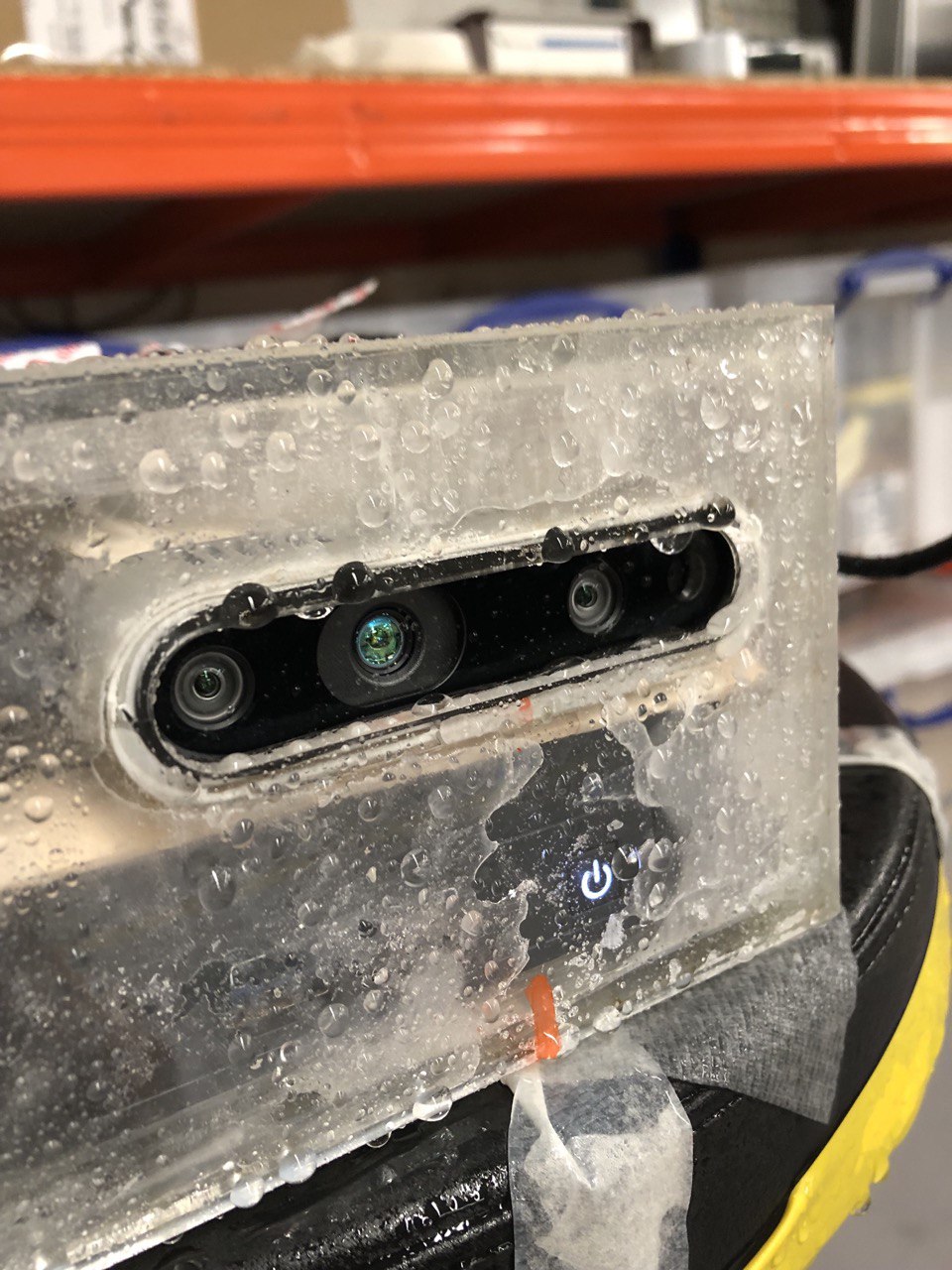}%
\label{box1}}
\hfil
\subfloat{\includegraphics[width=40mm]{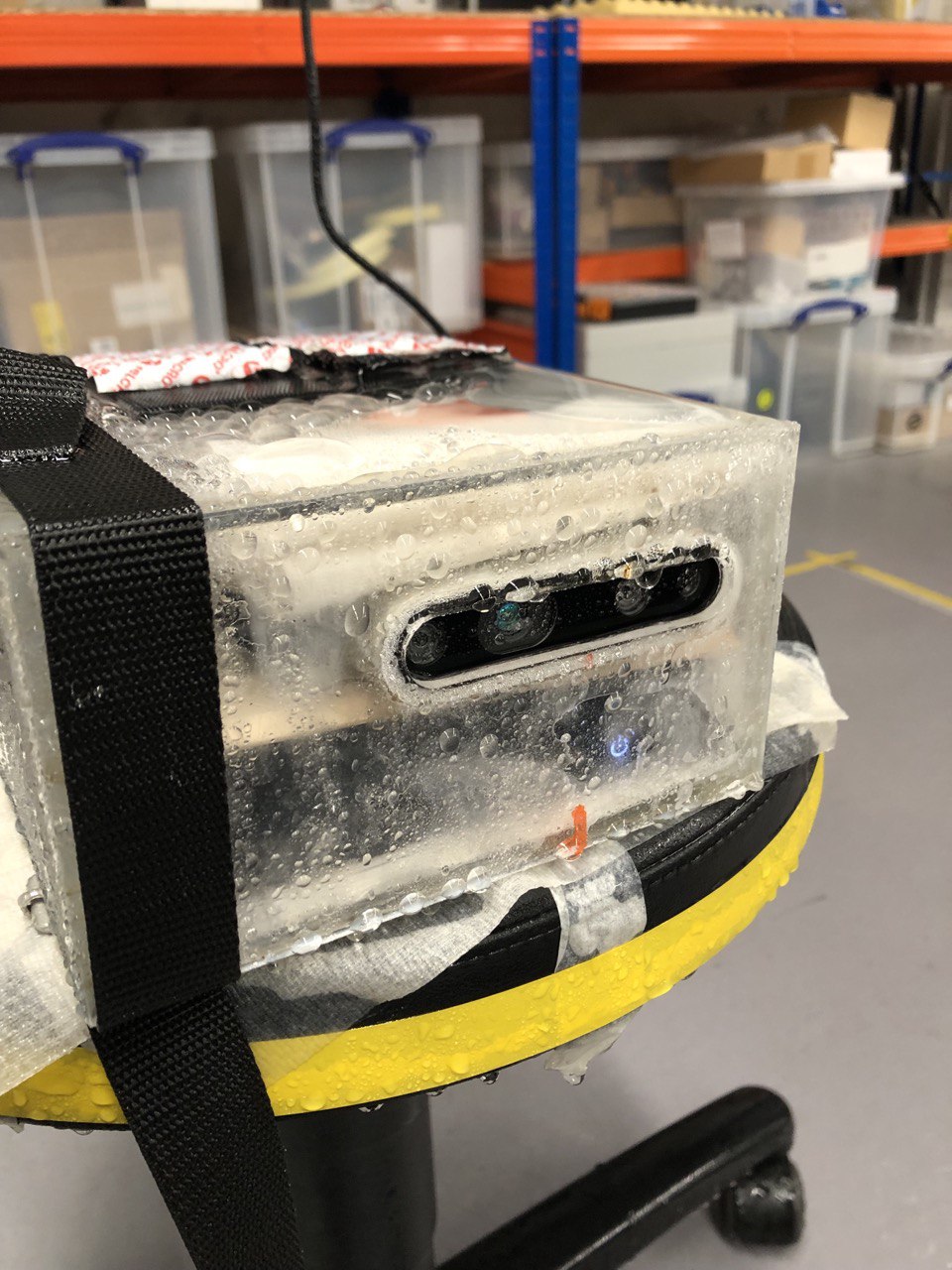}%
\label{box2}}
\caption{The water-resistant box that hosts the processing unit and the depth camera. Right hand side shows stool mounting for moving conditions (Sec.V).}
\label{box}
\end{figure}

Given that the electronics may become damaged by exposure to water, we designed and fabricated a water-resistant box to enclose and protect them, as shown in Figure~\ref{box}. 
The dimensions of the box are $20\times15\times10$ cm (length, width, height), and so it can easily host the processing unit, the depth camera, and the necessary cables. At the back of the box, there is a lid that permits the insertion and removal of the electronic devices; moreover, an IP68 nylon gland connects the device power supply to an external power source to further ensure water resistance. The manufacturing process has been conducted using a laser cutting machine to cut the box faces made of polymethyl methacrylate. Then, we composed the box with bi-component specific glue, silicon, and rubber seals to enhance water ingress protection.

\subsection{Visual Odometry Algorithm}
The processing unit is programmed with the ``VINS-Fusion" algorithm described in~\cite{qin2017vins, qin2018online, qin2019a, qin2019b}. It consists of an optimization-based multi-sensor state estimator that runs accurate simultaneous localization and mapping (SLAM) for autonomous navigation applications. The authors of VINS-Fusion developed the platform to support a variety of visual-inertial sensor types including mono camera and IMU, stereo cameras and IMU, and stereo cameras only. We specifically use the algorithm with stereo cameras only running on Intel Real Sense D435i\footnote{\url{https://github.com/HKUST-Aerial-Robotics/VINS-Fusion/tree/master/config/realsense_d435i}}. In this way, the inertial contribution is avoided, and we focus directly on the visual odometry.

\subsection{Experimental Environment and Settings}
The experiments were conducted in a controlled indoor laboratory, which is representative of a challenging and high-entropy scenario. Experiments were designed for two different navigation conditions, namely static and moving. In the static condition, the VO system is stationary in all axes and simulates a drone hovering. In the moving condition, the VO system follows a rectangular trajectory of size $140\times160$ cm with a fixed altitude. The constant height allows the simulation of the drone's navigation when it reaches the desired altitude. In this experiment, a specifically trained user moves the VO system over a stool to ensure a constant velocity of around 0.2 m/s (shown in Fig.~\ref{box}). We use a particularly low velocity to exclude its contribution to the analysis, thus focusing only on the rain effect and perturbation in the navigation system. Moreover, we simulate different rain conditions with a sprayer by changing its distance and inclination from the VO system, taking inspiration from~\cite{hasirlioglu2019general}. 
In particular, we use the rain conditions shown in Table~\ref{rain_conditions}. In addition, we conducted experiments in clear conditions without simulated rain to have a reference for the other experiments.
Overall, the slanting rain scenario has been designed to show and analyse the effect of raindrops directly on the camera lens. This scenario is the most common, as it is present during navigation at medium/high velocities. In contrast, the vertical rain scenario does not directly show the raindrop effect on the camera lens but happens as a small raindrop aggregation. This scenario reflects hovering or navigation at low velocities. Furthermore, we measured the time required to recover from the slanting rain condition (i.e., the most severe) needed to ensure a VO navigation with an average error below 30 cm, thus ensuring an acceptable error that does not lead to hazardous situations, especially in higher altitude flights.

\begin{table}[]
\centering
\caption{The different rain conditions used during the experiments. The distance (i.e., ``Dist.") shows the length between the sprayer and the box to simulate different rain intensities. The inclination is w.r.t. the vertical axes in front/side of the camera to avoid the visual perturbation on the camera scene by the operator.}
\label{rain_conditions}
\begin{tabular}{ccc}
\toprule
\textbf{Rain Conditions} & \textbf{Slanting}                                                                                          & \textbf{Vertical}                                                                                        \\ \midrule
\textbf{Heavy}           & \begin{tabular}[c]{@{}l@{}}Dist. \textless \space10 cm\\ Inclination $\sim$ 30°\end{tabular}                   & \begin{tabular}[c]{@{}l@{}}Dist. \textless \space 10 cm\\ Inclination $\sim$ 0°\end{tabular}                  \\
\textbf{Medium}          & \begin{tabular}[c]{@{}l@{}}10 cm \textless \space Dist.  \textless \space 20 cm \\ Inclination $\sim$ 30°\end{tabular} & \begin{tabular}[c]{@{}l@{}}10 cm \textless \space Dist.  \textless \space 20 cm \\ Inclination $\sim$ 0°\end{tabular} \\
\textbf{Low}             & \begin{tabular}[c]{@{}l@{}}Dist. \textgreater \space 30 cm\\ Inclination $\sim$ 30°\end{tabular}                & \begin{tabular}[c]{@{}l@{}}Dist. \textgreater \space 30 cm\\ Inclination $\sim$ 0°\end{tabular}               \\ \bottomrule
\end{tabular}
\end{table}

All the experiments have been repeated 30 times with a view to ensuring statistical soundness. The quantity of water released during each experiment is constant and consists of 1.8 ml/s, which is equivalent to 2.4 sprays/s (this rate is due to the natural and continuous spray rate of an average user).

\section{Dataset and DNN Development}
\subsection{Dataset} \label{dataset}
\begin{figure}[]
  \centering
  \begin{subfigure}[b]{0.45\linewidth}
    \includegraphics[width=\linewidth]{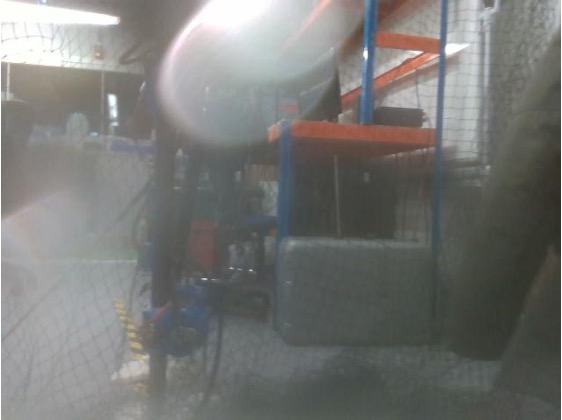}
     \caption{Slanting heavy rain.}
  \end{subfigure}
  \begin{subfigure}[b]{0.45\linewidth}
    \includegraphics[width=\linewidth]{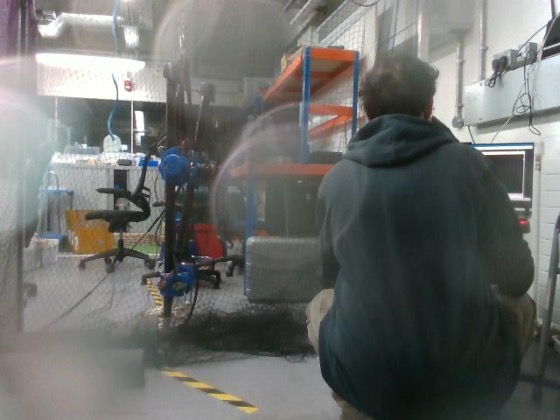}
    \caption{Slanting medium rain.}
  \end{subfigure}
  \begin{subfigure}[b]{0.45\linewidth}
    \includegraphics[width=\linewidth]{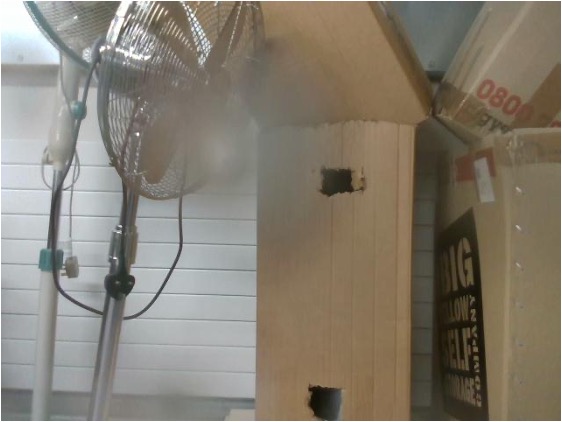}
    \caption{Slanting low rain.}
  \end{subfigure}
  \begin{subfigure}[b]{0.45\linewidth}
    \includegraphics[width=\linewidth]{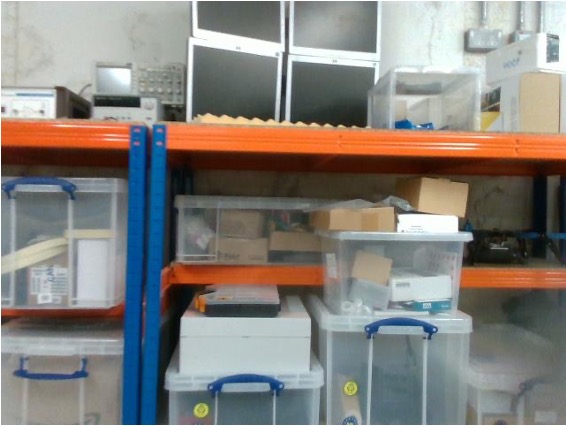}
    \caption{Vertical heavy rain.}
  \end{subfigure}
  \begin{subfigure}[b]{0.45\linewidth}
    \includegraphics[width=\linewidth]{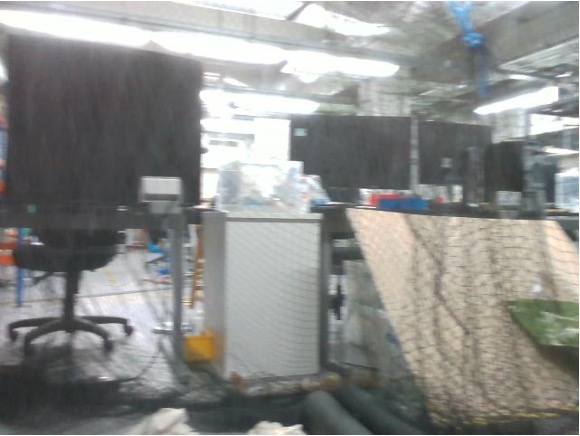}
    \caption{Vertical medium rain.}
  \end{subfigure}
  \begin{subfigure}[b]{0.45\linewidth}
    \includegraphics[width=\linewidth]{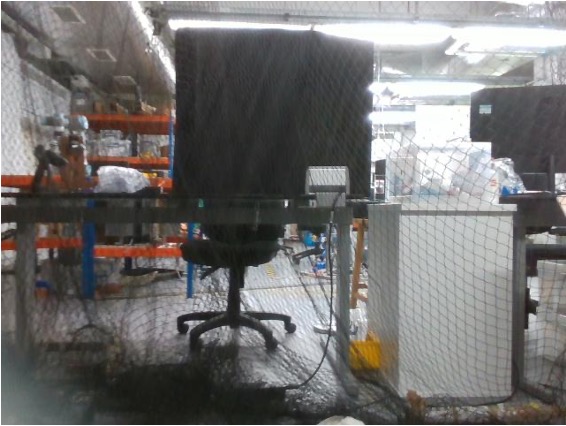}
    \caption{Vertical low rain.}
  \end{subfigure}
  \caption{Example of dataset images for each class of rain.}
  \label{fig:dataset_images}
\end{figure}

During the experiments presented in Section~\ref{expsetup}, we collected raw color images to construct a dataset representing the different rain conditions analysed in this study following~\cite{paullada2021data}. This data can be useful to develop a classification system that can understand the external condition and then act accordingly. The dataset consists of 7 classes of 48k images per class, namely ``Clear", ``Slanting Heavy Rain", ``Vertical Heavy Rain", ``Slanting Medium Rain", ``Vertical Medium Rain", ``Slanting Low Rain", and ``Vertical Low Rain". Figure~\ref{fig:dataset_images} shows several examples of images from the dataset. Overall, the dataset is composed of around 336k images and is almost 12 GB (compressed) in total, where 80\% is used for training, 10\% for validation and the remaining 10\% is used for testing.


\subsection{Deep Neural Network Specifications}
\label{DNN}
Three different DNNs have been selected and trained with the dataset developed in Section~\ref{dataset}. We use state-of-the-art architectures, namely MobileNetV2~\cite{sandler2018mobilenetv2} (alpha parameter equal to 0.35), MobileNetV3 small~\cite{howard2019searching} (alpha parameter equal to 0.35), and SqueezeNet~\cite{iandola2016squeezenet} as they show an optimal trade-off between performance and computational complexity~\cite{albanese2022tiny, albanese2023industrial} for similar mobile and/or constrained deployment contexts.
The DNNs are trained on an NVIDIA GeForce RTX 4090 GPU with the following hyperparameters:
\begin{itemize}
    \item 100 epochs
    \item Image shape $224\times224\times3$
    \item SGD optimizer
    \item Batch size 64
    \item Polynomial decay learning rate from $10^{-1}$ to $10^{-3}$ with $10^4$ decay steps and square root function (i.e., power equal to 0.5)
\end{itemize}

\begin{table}[]
\centering
\caption{Number of parameters and memory footprint of the three architectures presented.}
\label{npara}
\begin{tabular}{ccc}
\toprule
\textbf{Architecture}                                                                      & \multicolumn{1}{c}{\textbf{Number of Parameters}} & \multicolumn{1}{c}{\textbf{Memory Footprint (MB)}} \\ \midrule
\textbf{MobileNetV2}                                                                      & 419175                                            & 1.70                                               \\
\multicolumn{1}{c}{\textbf{\begin{tabular}[c]{@{}c@{}}MobileNetV3 \\ Small\end{tabular}}} & 336855                                            & 1.28                                               \\
\textbf{SqueezeNet}                                                                        & 774503                                            & 2.95                                               \\ \bottomrule
\end{tabular}
\end{table}

Table~\ref{npara} summarizes the number of parameters and the memory footprint of the three architectures. Even though they have a deep structure, they present a low memory footprint because of the innovative blocks that compose them.
MobileNetV3 Small uses significantly fewer parameters compared to other models, resulting in lower memory usage of only 1.28 MB. This is 24\% less than MobileNetV2 and less than half that of SqueezeNet. This reduction in parameters is achieved through a combination of design and architectural optimizations in MobileNetV3 Small.

\section{Experimental Results \& Analysis}
In the first part of this section, we present the experimental results analysed with the setup presented in Section~\ref{expsetup}. We analyse the experiments in static and moving conditions. First, we use the standard deviation to provide the error of the path estimation of the VO system. In the second, we use the root mean square error (RMSE) to provide, on average, the error on the path estimation of the VO system by using the clear condition as the reference.

We also report the time required in the rain scenario to ensure a return to navigation accuracy with an acceptable error. We define `Restoring Time' as the duration needed to reduce the error to below 30 cm after encountering adverse rain conditions. In our experiments, the system remained operational during and after the rain emulation. Once the rain ceased, the error remained above 30 cm for some tens of seconds before stabilizing within acceptable limits.
The longest Restoring Time, approximately 33 seconds, was observed during slanting rain due to the increased likelihood and intensity of lens wetting. Concerning DNN performance, we analyze the confusion matrix of the 7-class classifier, as provided by the source code, to calculate average accuracy, precision, recall, and F1-score. These metrics are used to identify the best-performing DNN.


\subsection{VO System in Static and Moving Conditions}
\label{VOresults}
The VO system developed in Section~\ref{expsetup} has been tested in static and moving conditions. For each condition, we evaluated different rain intensities and modalities, namely {\em slanting} heavy rain, vertical heavy rain, slanting medium rain, vertical medium rain, slanting low rain, and vertical low rain (as shown in Table~\ref{expsetup}).

\subsubsection{Static}
In this experiment, 
the VO system is completely stationary at the same point for the duration of the trials.
\begin{figure*}[]
  \centering
  \begin{subfigure}[t]{0.31\linewidth}
    \includegraphics[width=\linewidth]{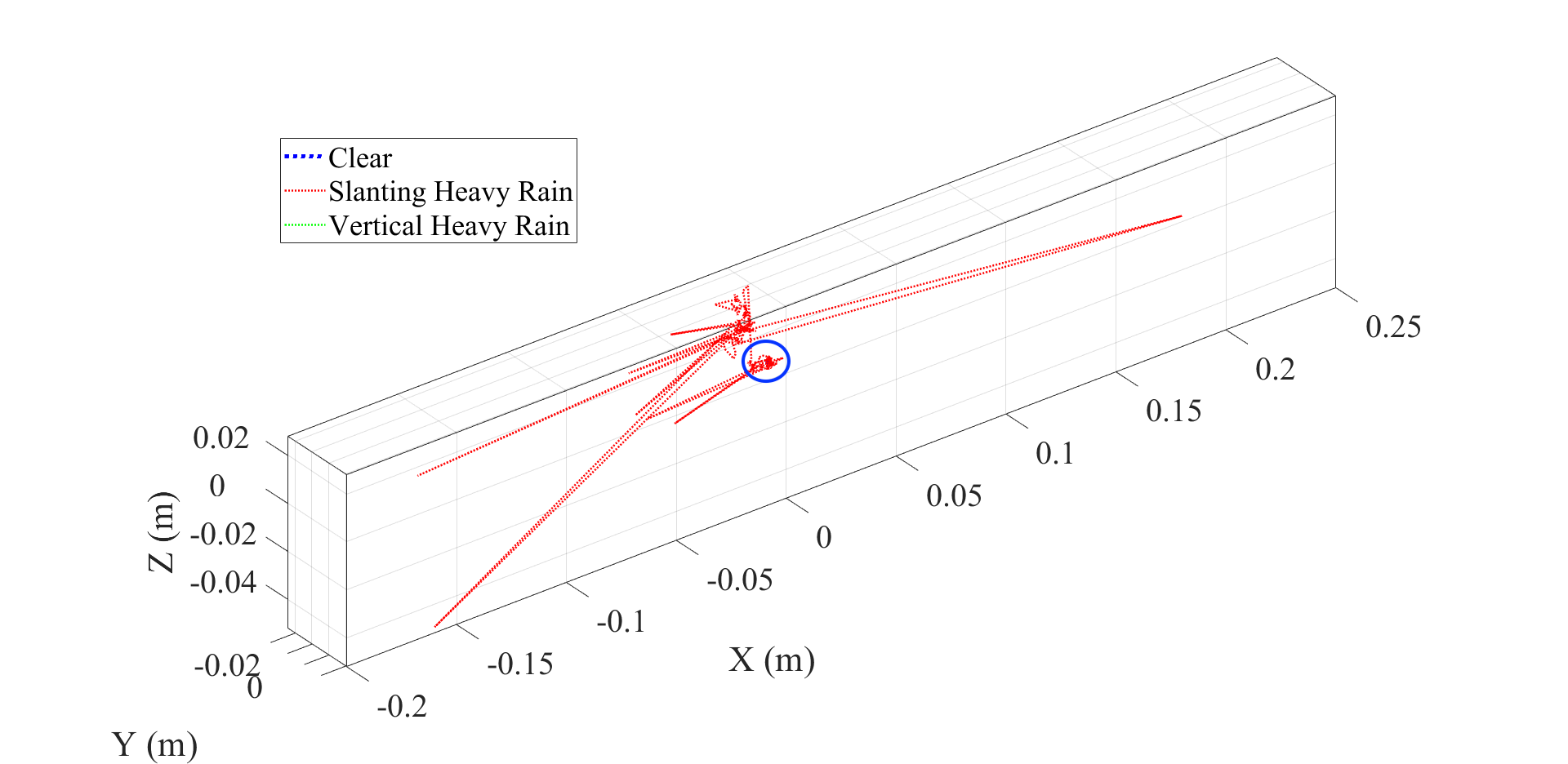}
     \caption{Trajectory estimation of slanting and vertical heavy rain. The blue circle highlights the clear and vertical rain trajectories.}
     \label{traj_hr}
  \end{subfigure}
  \begin{subfigure}[t]{0.31\linewidth}
    \includegraphics[width=\linewidth]{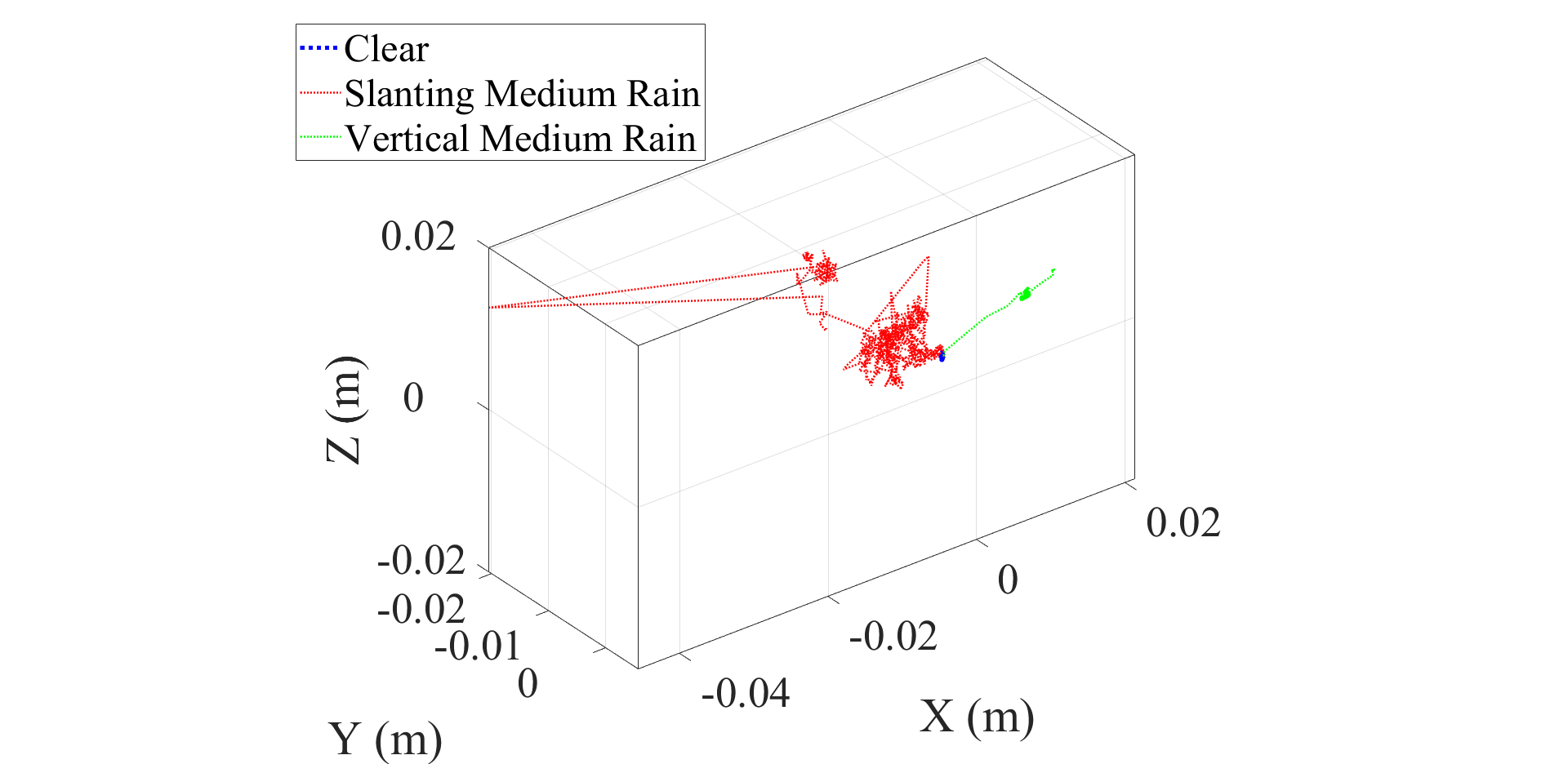}
    \caption{Trajectory estimation of slanting and vertical medium rain.}
  \end{subfigure}
  \begin{subfigure}[t]{0.31\linewidth}
    \includegraphics[width=\linewidth]{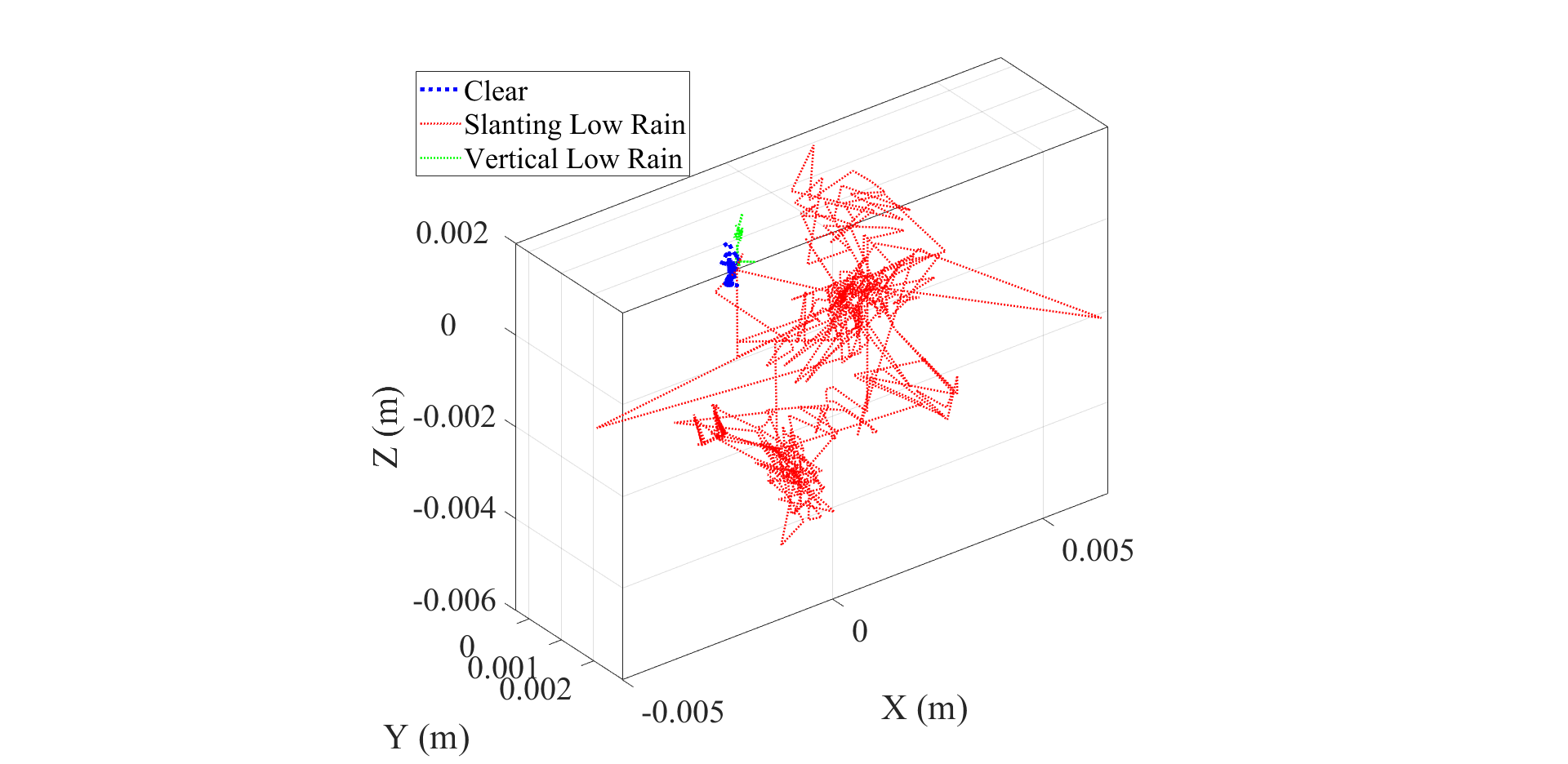}
    \caption{Trajectory estimation of slanting and vertical low rain.}
  \end{subfigure}
  \begin{subfigure}[t]{0.23\linewidth}
    \includegraphics[width=\linewidth]{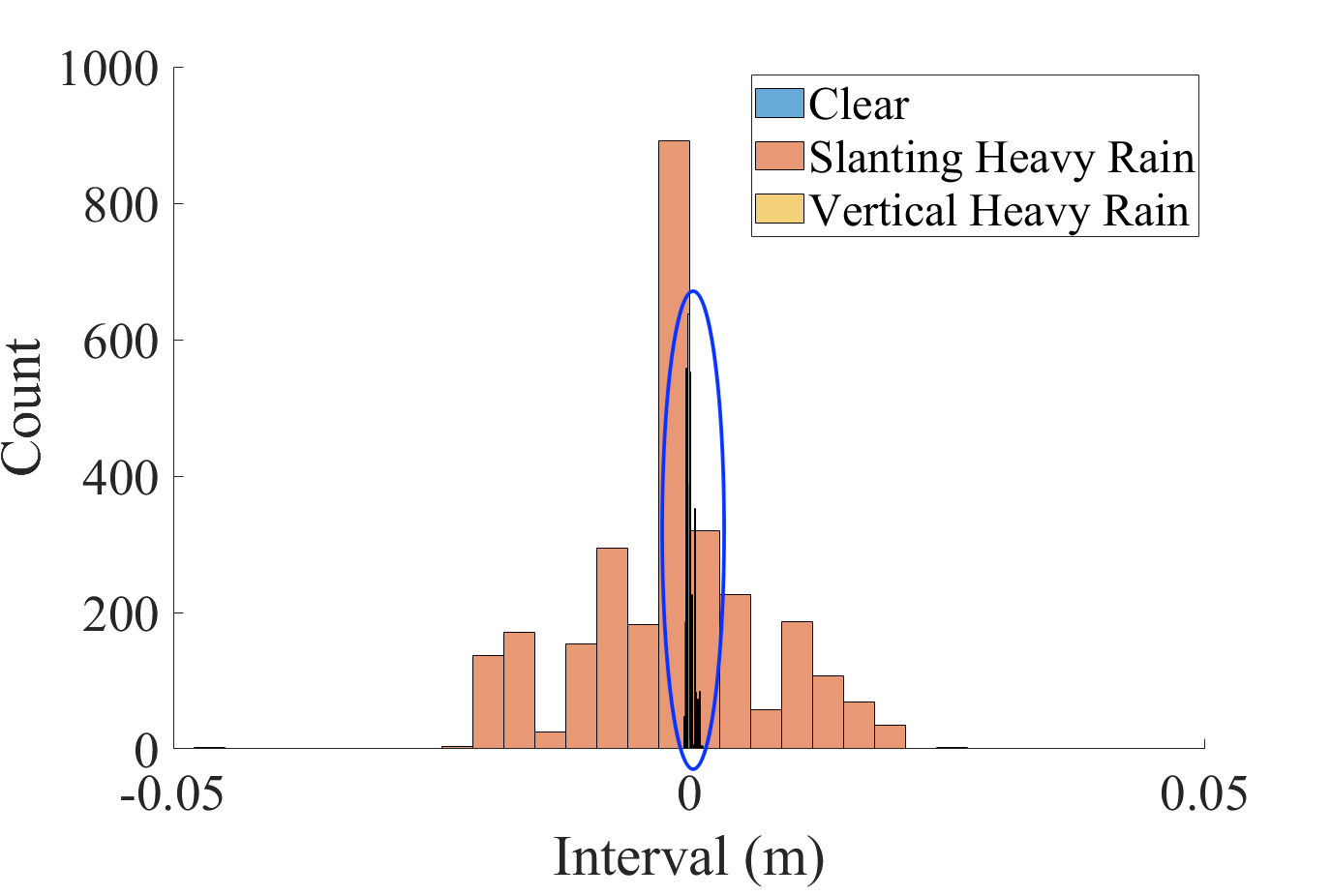}
    \caption{Data distribution of slanting and vertical heavy rain. The blue circle highlights the clear and vertical rain data distributions.}
    \label{data_hr}
  \end{subfigure}
  \begin{subfigure}[t]{0.31\linewidth}
    \includegraphics[width=\linewidth]{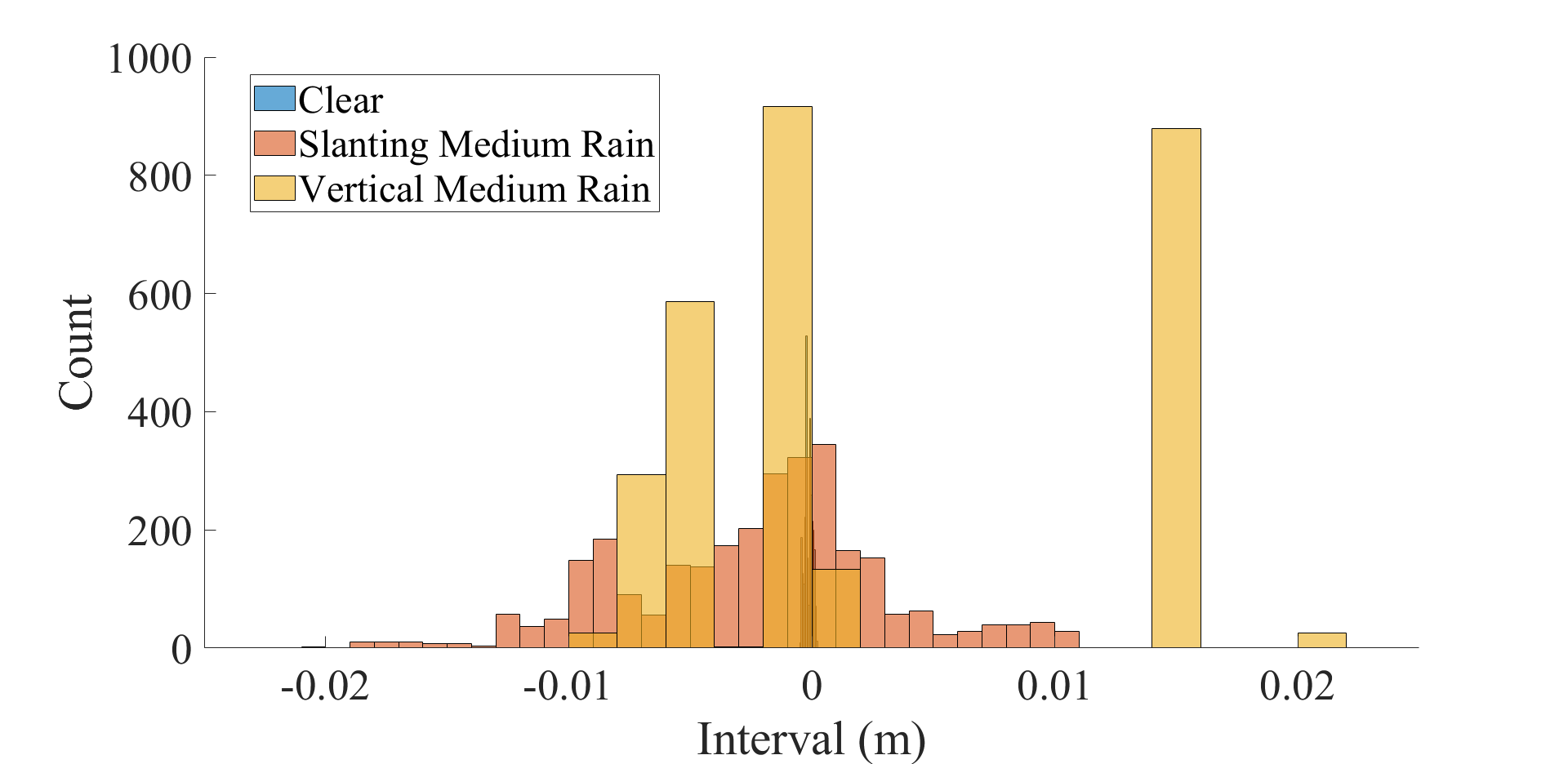}
    \caption{Data distribution of slanting and vertical medium rain.}
  \end{subfigure}
  \begin{subfigure}[t]{0.31\linewidth}
    \includegraphics[width=\linewidth]{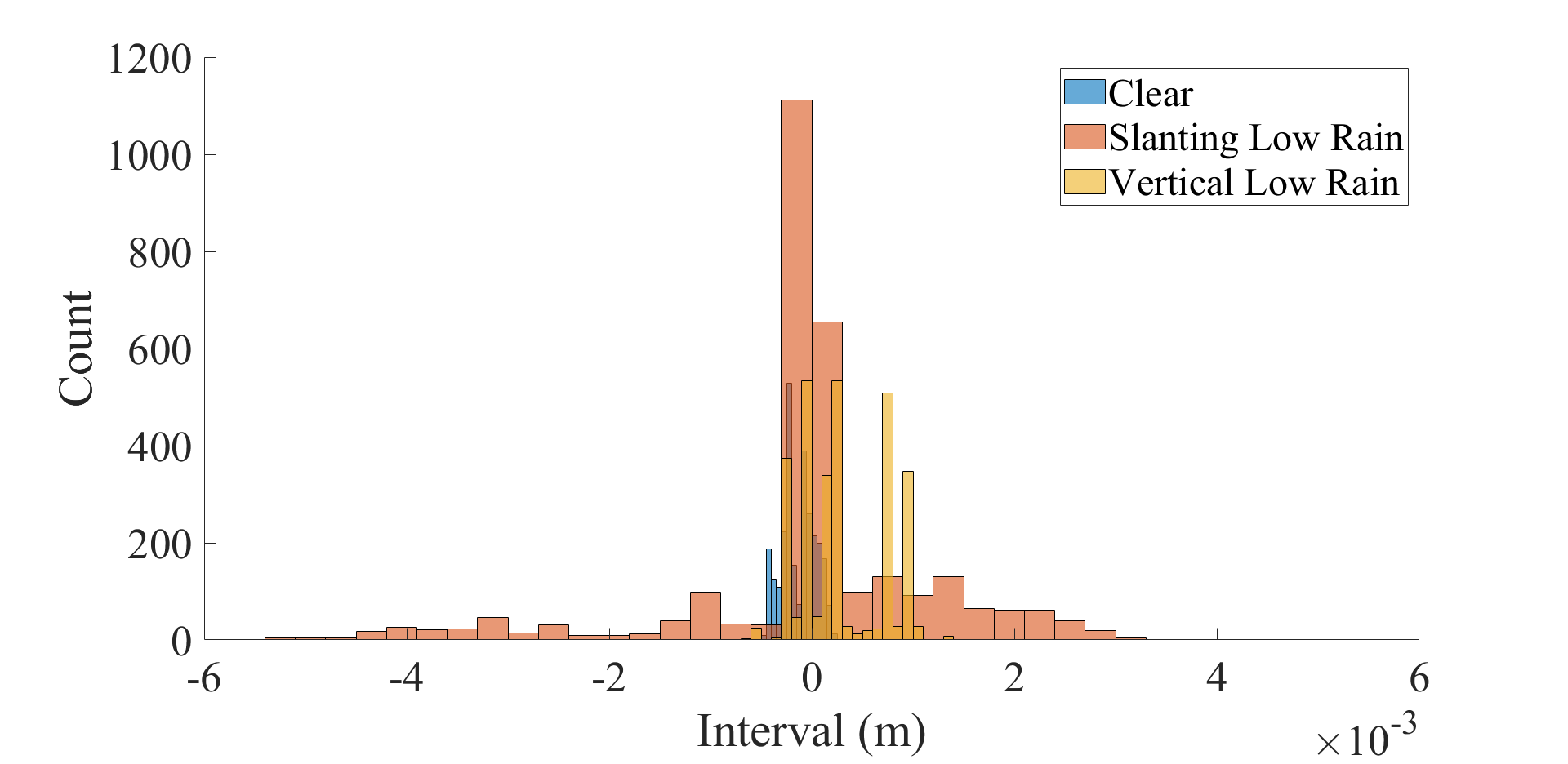}
    \caption{Data distribution of slanting and vertical low rain.}
  \end{subfigure}
  \caption{Trajectory estimation and data distribution of the static scenario under the different rain conditions.}
  \label{fig:static_exp}
\end{figure*}
Figure~\ref{fig:static_exp} shows a comparison of the trajectory estimation and data distribution of the experiments conducted in a static scenario under different rain conditions. Furthermore, Table~\ref{static} summarizes the error computed in terms of standard deviation as the system is fixed and the real position is 0 m in all axes. In this way, it is possible to evaluate the performance degradation starting from the optimal condition (i.e., clear) to the worst-case scenario (i.e., slanting heavy rain). Figures~\ref{traj_hr} and~\ref{data_hr} reveal the severity of the slanting heavy rain scenario confirmed by a considerable drift in Table~\ref{static}. On the other hand, the ``Vertical Low Rain" scenario is almost comparable to the ``Clear" one, meaning that the rain perturbation can be negligible.

\begin{table}[]
\centering
\caption{Error computed of the experiments in static condition in terms of standard deviation over the three axes x, y, and z. The worst case scenario is ``Slanting Heavy Rain" (highlighted in red), while the best case scenario is ``Vertical Low Rain" (highlighted in green).}
\label{static}
\begin{tabular}{cc}
\toprule
\textbf{Condition (Static)}                               & \multicolumn{1}{c}{\textbf{Error ($\sigma{_x}$, $\sigma{_y}$, $\sigma{_z}$)(mm)}} \\ \midrule
\textbf{Clear (Reference)}                & 0.05, 0.09, 0.2                                           \\
\textcolor{red}{\textbf{Slanting Heavy Rain}}                     & 10.5, 6.2, 7.9                                            \\
\textbf{Vertical Heavy Rain} & 0.1, 0.07, 0.3                                            \\
\textbf{Slanting Medium Rain}                    & 3.6, 5.1, 4.5                                             \\
\textbf{Vertical Medium Rain}                    & 3.5, 1.4, 0.4                                             \\
\textbf{Slanting Low Rain}                       & 1.0, 0.6, 1.4                                             \\
\textcolor{green}{\textbf{Vertical Low Rain}}                       & 0.09, 0.09, 0.2                                           \\ \bottomrule
\end{tabular}
\end{table}

\subsubsection{Moving}
\begin{figure*}[]
  \centering
  \begin{subfigure}[b]{0.31\linewidth}
    \includegraphics[width=\linewidth]{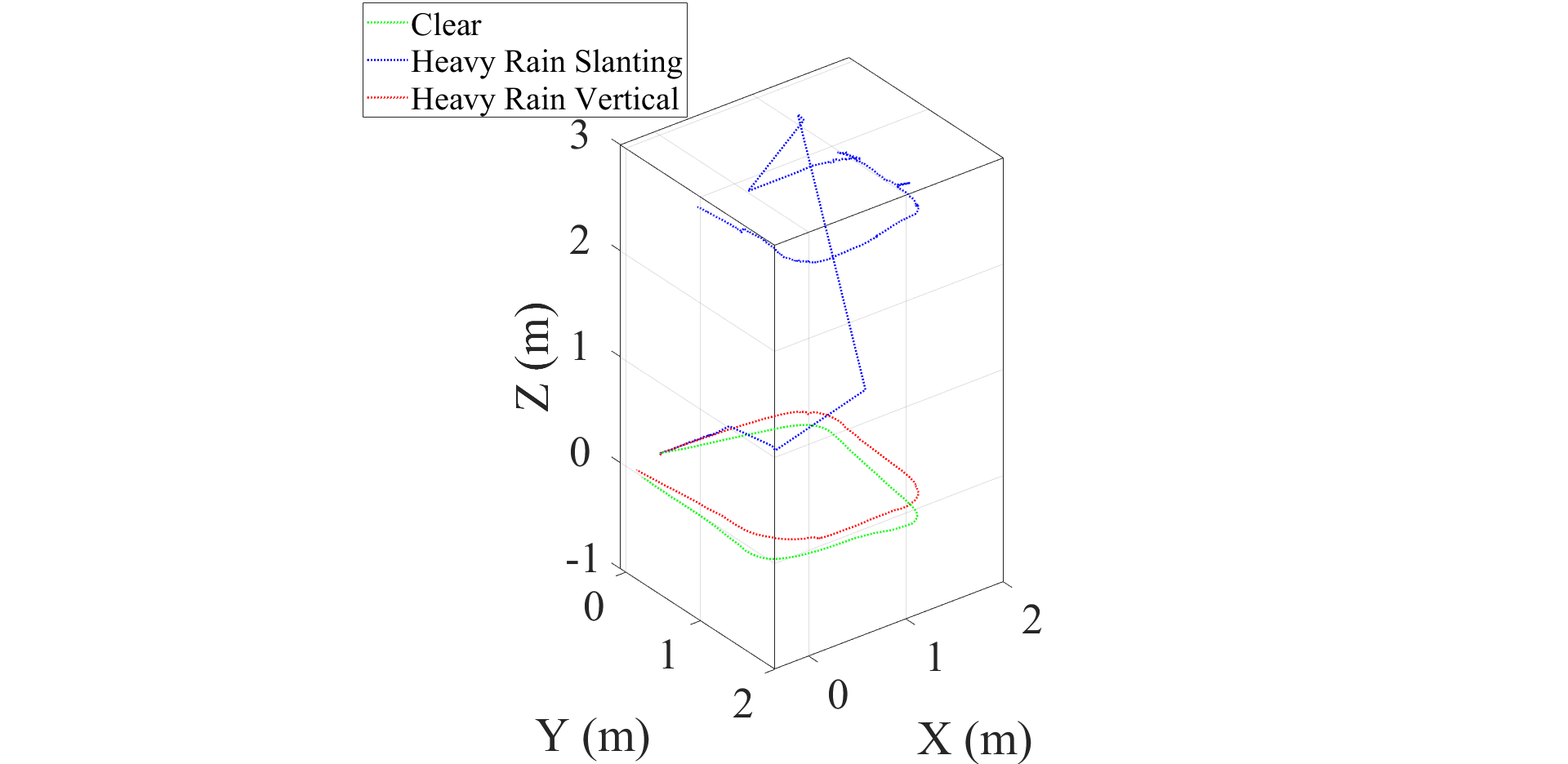}
     \caption{Trajectory estimation of slanting and vertical heavy rain.}
     \label{move_hr}
  \end{subfigure}
  \begin{subfigure}[b]{0.31\linewidth}
    \includegraphics[width=\linewidth]{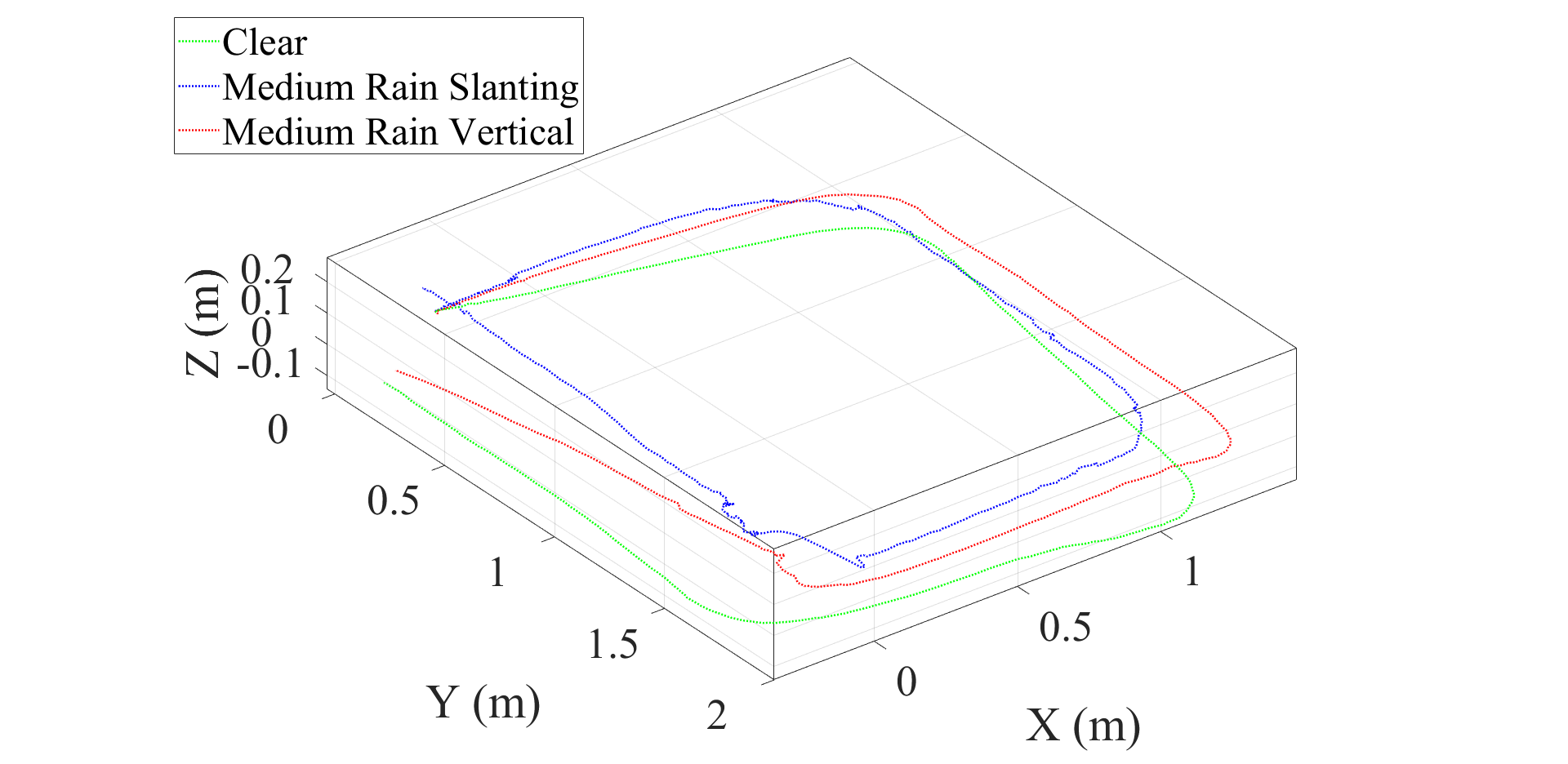}
    \caption{Trajectory estimation of slanting and vertical medium rain.}
  \end{subfigure}
  \begin{subfigure}[b]{0.31\linewidth}
    \includegraphics[width=\linewidth]{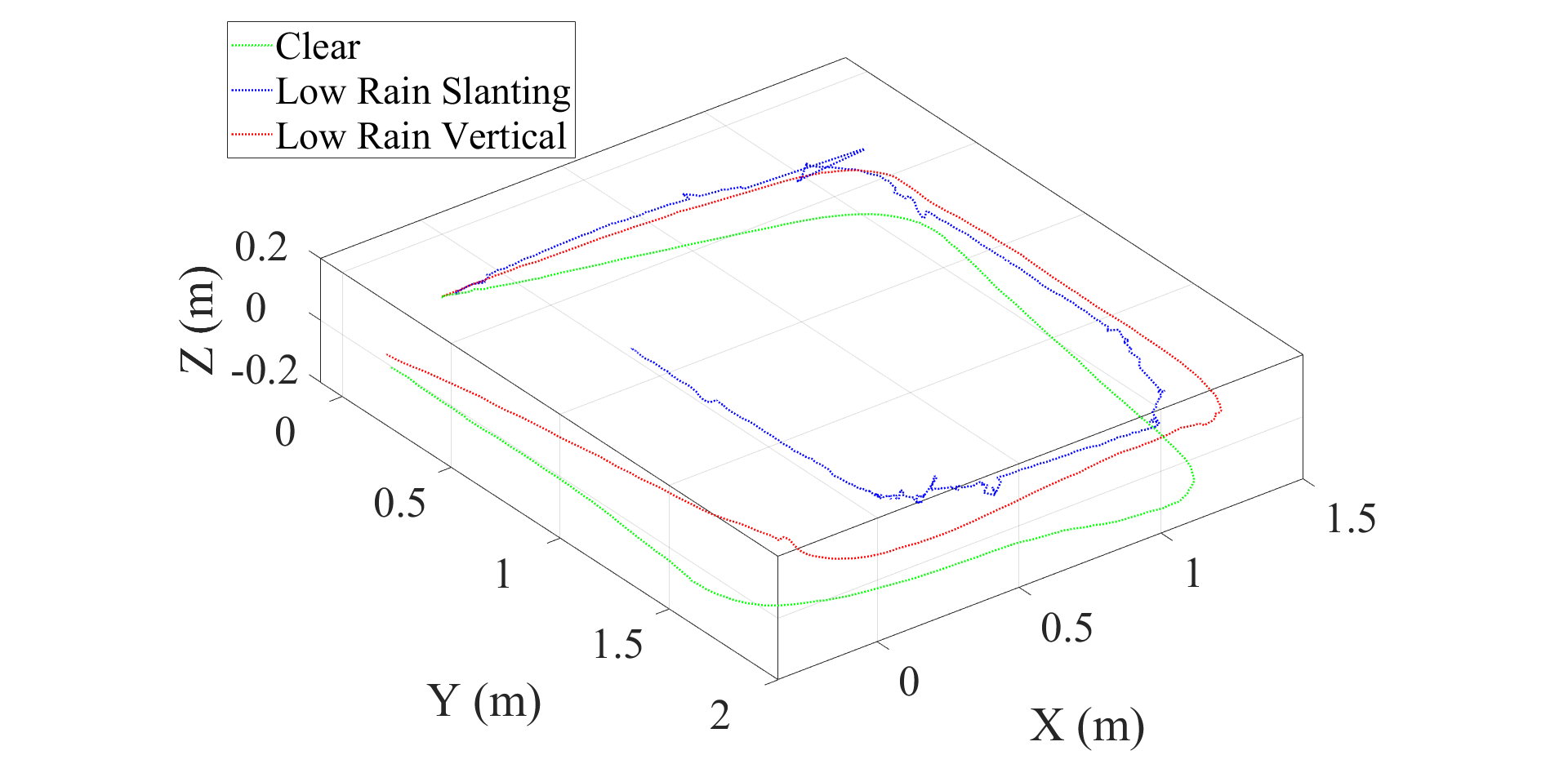}
    \caption{Trajectory estimation of slanting and vertical low rain.}
  \end{subfigure}
  \caption{Trajectory estimation and data distribution of the moving scenario under the different rain conditions.}
  \label{fig:moving_exp}
\end{figure*}

In this experiment, we analyse and compare the effect of the different rain conditions of Table~\ref{expsetup} in a moving scenario (i.e., the VO system following a rectangular trajectory). Figure~\ref{fig:moving_exp} shows a comparison of the trajectory estimation under the different rain conditions broken down depending on the rain intensity (i.e., heavy, medium, and low rain). Furthermore, Table~\ref{moving} summarizes the RMSE computed by using the clear condition as the reference. This analysis confirms the severity of the slanting heavy rain scenario, which introduces an unacceptable drift in all axes, especially in the vertical one (Figure~\ref{move_hr}). On the other hand, the vertical rain scenarios are almost comparable and less alarming as they show drift in the order of dozens of centimetres.

\begin{table}[]
\centering
\caption{RMSE over the three axes and Restoring Time of the moving scenario. The worst-case scenario is ``Slanting Heavy Rain” (highlighted in red), while the best-case scenarios are ``Vertical Medium Rain” and ``Vertical Low Rain” (highlighted in green).}
\label{moving}
\begin{tabular}{ccc}
\toprule
\textbf{Condition (Moving)}   & \multicolumn{1}{c}{\textbf{RMSE (x, y, z)(m)}} & \textbf{Restoring Time (s)} \\ \midrule
\textcolor{red}{\textbf{Slanting Heavy Rain}}  & 1.3, 0.9, 2.5                                  & 32.9                        \\
\textbf{Vertical Heavy Rain}  & 0.4, 0.4, 0.09                                 & NA                          \\
\textbf{Slanting Medium Rain} & 0.5, 0.5, 0.3                                  & 20.1                        \\
\textcolor{green}{\textbf{Vertical Medium Rain}} & 0.3, 0.3, 0.08                                 & NA                          \\
\textbf{Slanting Low Rain}    & 0.8, 0.9, 0.4                                  & 13.34                       \\
\textcolor{green}{\textbf{Vertical Low Rain}}    & 0.3, 0.4, 0.07                                 & NA                          \\ \bottomrule
\end{tabular}
\end{table}

\subsection{DNN Test}
\label{DNNtest}
We test the DNNs developed in Section~\ref{DNN} by using their confusion matrix. In particular, each architecture is evaluated by considering the precision, recall, and f1-score of each class as shown in Tables~\ref{mbv2},~\ref{mbv3small}, and~\ref{snet}. Metrics are computed with the test set of the dataset presented in Section~\ref{dataset}, which consists of 10\% of the total dataset size. The three architectures perform well in all classes, especially for the clear and slanting heavy rain scenario. This is because they are the two extremes of the conditions studied, meaning that the system can easily discriminate these two scenarios. However, the ``Vertical Low Rain" class presents the lowest recall (highlighted in red in Tables~\ref{mbv2},~\ref{mbv3small}, and~\ref{snet}), meaning the system recognizes many false negatives. We expected such behaviour as the perturbation introduced by a vertical low rain scenario is very small and almost comparable to the clear scenario (as analysed in Section~\ref{VOresults}). Nonetheless, considering UAV autonomous navigation, a false negative in a vertical low rain scenario does not present a hazard during navigation, thus making it acceptable.

Moreover, we provide the overall accuracy, precision, recall, and f-score in Table~\ref{perf}. The computational latency of the three classifiers on the Intel NUC 11 is shown in Table~\ref{latency}. The performance of the three DNNs is comparable, ensuring an accuracy, precision, recall, and f1-score over 90\%. On the other hand, MobileNet V3 Small presents the lowest memory footprint (Table~\ref{npara}) and the lowest classification latency of around 10 ms (Table~\ref{latency}), that means achieving slightly more than 93 FPS, a marked improvement over other methods.

The update rate or control frequency of an Electronic Speed Controller (ESC) for UAVs is generally in the range of 50 to 500~Hz. This means that the ESC receives and processes control signals from the flight controller approximately 50 to 500 times per second.
This update rate allows the ESC to adjust the motor speed rapidly, ensuring precise control of the UAV’s propulsion system~\cite{ESC}. The latency of 10.7ms (93~FPS) achieved by our work, permits taking immediate decisions and intervening promptly on drone control with cycle-accuracy.
For this reason, MobileNetV3 Small is a suitable choice for near real-time applications on edge devices such as small-size UAVs.

\begin{table}[]
\centering
\caption{Performance of each class of MobileNetV2. The worst performance is the recall in the ``Vertical Low Rain" condition highlighted in red.}
\label{mbv2}
\begin{tabular}{cccc}
\toprule
\textbf{MobileNet V2}         & \textbf{Precision} & \multicolumn{1}{c}{\textbf{Recall}} & \multicolumn{1}{c}{\textbf{F1-score}} \\ \midrule
\textbf{Clear}                & 0.99               & 0.99                                & 0.99                                  \\
\textbf{Slanting Heavy Rain}  & 0.99               & 0.98                                & 0.99                                  \\
\textbf{Vertical Heavy Rain}  & 0.78               & 0.99                                & 0.88                                  \\
\textbf{Slanting Medium Rain} & 0.97               & 0.79                                & 0.87                                  \\
\textbf{Vertical Medium Rain} & 0.76               & 0.99                                & 0.86                                  \\
\textbf{Slanting Low Rain}    & 0.96               & 0.91                                & 0.93                                  \\
\textbf{Vertical Low Rain}    & 0.93               & \textcolor{red}{0.68}                                & 0.79                                  \\ \bottomrule
\end{tabular}
\end{table}

\begin{table}[]
\centering
\caption{Performance of each class of MobileNetV3 Small. The worst performance is the recall in the ``Vertical Low Rain" condition highlighted in red.}
\label{mbv3small}
\begin{tabular}{cccc}
\toprule
\textbf{MobileNet V3 Small}         & \textbf{Precision} & \multicolumn{1}{c}{\textbf{Recall}} & \multicolumn{1}{c}{\textbf{F1-score}} \\ \midrule
\textbf{Clear}                & 0.97               & 0.99                                & 0.98                                  \\
\textbf{Slanting Heavy Rain}  & 0.98               & 0.99                                & 0.98                                  \\
\textbf{Vertical Heavy Rain}  & 0.86               & 0.95                                & 0.90                                  \\
\textbf{Slanting Medium Rain} & 0.97               & 0.88                                & 0.92                                  \\
\textbf{Vertical Medium Rain} & 0.75               & 0.97                                & 0.84                                  \\
\textbf{Slanting Low Rain}    & 0.93               & 0.91                                & 0.92                                  \\
\textbf{Vertical Low Rain}    & 0.89               & \textcolor{red}{0.62}                                & 0.73                                  \\ \bottomrule
\end{tabular}
\end{table}

\begin{table}[]
\centering
\caption{Performance of each class of SqueezeNet. The worst performance is the recall in the ``Vertical Low Rain" condition highlighted in red.}
\label{snet}
\begin{tabular}{cccc}
\toprule
\textbf{SqueezeNet}           & \textbf{Precision} & \multicolumn{1}{c}{\textbf{Recall}} & \multicolumn{1}{c}{\textbf{F1-score}} \\ \midrule
\textbf{Clear}                & 0.97               & 0.99                                & 0.98                                  \\
\textbf{Slanting Heavy Rain}  & 0.98               & 1.00                                & 0.99                                  \\
\textbf{Vertical Heavy Rain}  & 0.95               & 1.00                                & 0.97                                  \\
\textbf{Slanting Medium Rain} & 0.94               & 0.82                                & 0.88                                  \\
\textbf{Vertical Medium Rain} & 0.76               & 0.98                                & 0.86                                  \\
\textbf{Slanting Low Rain}    & 0.98               & 0.88                                & 0.93                                  \\
\textbf{Vertical Low Rain}    & 0.83               & \textcolor{red}{0.72}                                & 0.77                                  \\ \bottomrule
\end{tabular}
\end{table}

\begin{table}[]
\centering
\caption{Performance of the three architectures.}
\label{perf}
\begin{tabular}{ccccc}
\toprule
\textbf{Architecture}       & \multicolumn{1}{c}{\textbf{Accuracy}} & \textbf{Precision} & \multicolumn{1}{c}{\textbf{Recall}} & \multicolumn{1}{c}{\textbf{F1-score}} \\ \midrule
\textbf{MobileNetV2}       & 0.90                                  & 0.91               & 0.91                                & 0.90                                  \\
\textbf{MobileNetV3 Small} & 0.90                                    & 0.91                 & 0.90                                  & 0.90                                    \\
\textbf{SqueezeNet}         & 0.91                                  & 0.92               & 0.91                                & 0.91                                  \\ \bottomrule
\end{tabular}
\end{table}


\begin{table*}[t]
\centering
\caption{Computational latency in seconds of the three classifiers on the Intel NUC 11.}
{\label{latency}
\begin{tabular}{cccccc}
\toprule
\textbf{Classifiers}     & \textbf{FPS}   & \textbf{Avg. (ms)} & \textbf{Var.($\times10^{-5}$) ($s^{2}$)} & \multicolumn{1}{c}{\textbf{Max. (s)}} & \multicolumn{1}{c}{\textbf{Min. (s)}} \\ \midrule
\textbf{MobileNetV2}     &14.16  & 70.6                                  & 3.2123               & 0.0944                                & 0.0651                                  \\
\multicolumn{1}{c}{\textbf{\begin{tabular}[c]{@{}c@{}}MobileNetV3 \\ Small\end{tabular}}} & 93.46  & 10.7                                    & 0.2062                 & 0.016.7                                  & 0.0087                                    \\
\textbf{SqueezeNet}       & 1.9  & 50.97                                  & 73.476               & 0.6385                                & 0.4791                                  \\ \bottomrule
\end{tabular}}
\end{table*}

\section{Conclusion}
\label{conclusion}
Autonomous UAV navigation based on VO systems operating under variable or adverse weather conditions is an understudied topic in the literature. Given that UAV employment is increasing rapidly in IoT contexts, it is fundamental to develop solutions to mitigate potentially hazardous situations (e.g., navigation in the rain) and increase drone `flyability'.

In this paper, we investigated the behaviour of a VO system under different rainy conditions that might be encountered during autonomous UAV navigation. Intuitively, the analysis determined that the worst-case scenario is slanting heavy rain, which was demonstrated to introduce an unacceptable error (i.e., from 1 m to 2.5 m) in the path estimation, thereby making the navigation dangerous and unreliable. On the other hand, the other ``slanting rain" scenarios (i.e., slanting medium rain and slanting low rain) present a higher error compared with the vertical rain conditions, meaning that they have to be evaluated depending on the application requirements.

We demonstrate a basis for the development of solutions to mitigate the effects of rain and make the autonomous UAV systems more `aware' of dynamic \textit{in situ} environmental conditions. Having trained and compared models based on three candidate DNN architectures, we have demonstrated that an accuracy of around 90\% can be achieved in classifying the colour images used by the VO system across 7 classes: clear, slanting heavy rain, vertical heavy rain, slanting medium rain, vertical medium rain, slanting low rain, and vertical low rain. The best-performing architecture has reached a frame rate of 93 FPS, achieving a latency of $\sim$10~ms, which is sufficient to enable near real-time UAV control. 

Accordingly, it is possible to use this information and incorporate these techniques in the development of disturbance estimation approaches feeding into online flight controllers that may allow for specific counteractions to be taken by the UAV (e.g., switch to an alternative navigation system, change the navigation path, land, etc.) depending on the environmental condition. This may permit performance improvements for navigation in otherwise hazardous situations, e.g., by avoiding collisions and increasing the UAV's reliability and flyability. Taking an `Internet of Things' perspective, the ability of a UAV to determine the local environmental conditions in real time may also be of significant importance to the development of more accurate and timely localised weather forecasting methods through the provision of this data. This and the integration of precipitation-based disturbance estimation with autonomous UAV tracking and trajectory control systems are left for future work.

\balance
\bibliographystyle{IEEEtran}
\bibliography{biblio}

\vfill

\end{document}